\pdfoutput=1
\documentclass[runningheads]{llncs}
\usepackage[T1]{fontenc}
\usepackage{graphicx}
\usepackage{hyperref}
\usepackage{amsfonts}
\usepackage{xcolor}
\usepackage{booktabs}
\usepackage{amssymb}
\makeatletter
\renewenvironment{thebibliography}[1]
  {\section*{\refname}%
   \def\@biblabel##1{##1.}%
   \small
   \list{\@biblabel{\@arabic\c@enumiv}}%
        {\settowidth\labelwidth{\@biblabel{#1}}%
         \leftmargin\labelwidth
         \advance\leftmargin\labelsep
         \usecounter{enumiv}%
         \let\p@enumiv\@empty
         \renewcommand\theenumiv{\@arabic\c@enumiv}%
         \itemsep\z@
         \parsep\z@
         \topsep\z@
         \partopsep\z@}%
   \renewcommand\newblock{\hskip .11em \@plus.33em \@minus.07em}%
   \sloppy\clubpenalty4000\widowpenalty4000%
   \sfcode`\.=\@m}%
  {\def\@noitemerr{\@latex@warning{Empty `thebibliography' environment}}%
   \endlist}
\makeatother
\begin{document}
\title{TextileNet: Towards Zero-shot Text-style Segmentation of Manuscripts}
\titlerunning{TextileNet: Zero-shot Text-style Segmentation}
\author{%
Anguelos Nicolaou\inst{1}\orcidID{0000-0003-3818-8718}\thanks{Corresponding author.} \and
Antonella Ambrosio\inst{2}\orcidID{0000-0001-6944-2147} \and
Desiree Di Donato\inst{2} \and
Georg Vogeler\inst{1}\orcidID{0000-0002-1726-1712}%
}

\authorrunning{A. Nicolaou et al.}

\institute{%
University of Graz, Graz, Austria\\
\email{anguelos.nicolaou@gmail.com, georg.vogeler@uni-graz.at}
\and
Università degli Studi di Napoli Federico II, Naples, Italy\\
\email{aambrosio@unina.it, desireedidonato40@gmail.com}%
}
\maketitle
\begin{abstract}
Automatic writer identification systems have progressed remarkably in recent years, yet their deployment in archival paleography remains limited by the scarcity of labeled training data, open scribe sets, and degraded image quality.
We present TextileNet, a fully convolutional multi-task network trained exclusively on synthetic data to produce dense pixel-level texture embeddings, which we transfer zero-shot to historical manuscript analysis.
As an original contribution to evaluation methodology, we designed a paleographic visual quiz of 80 pair and triplet questions and administered it to a range from lay participants to senior paleographers under strict anonymity, establishing to our knowledge for the first time a human baseline for script-style discrimination on late medieval text.
We employ TextileNet embeddings to perform zero-shot retrieval on sub-word granularity for hand and gender identification.
Our experimental results help in building the credibility of TextileNet in the paleographic domain, but more than that demonstrate in experimental terms that the question of gender in handwriting needs to be treated with caution.
\footnote{source code repository available: \url{https://github.com/anguelos/textstyle}}

\end{abstract}

\section{Introduction}

\subsection{Writer identification methods}

Automatic writer identification has been an active field in document image analysis for several decades, driven by applications ranging from forensic document examination to the study of historical manuscripts.
Early approaches established the viability of local texture statistics and sparse coding as writer-specific descriptors~\cite{nicolaou2014local,nicolaou2015sparse,nicolaou2016visual}.
Even when deep learning dominated other domains, writer identification state-of-the-art was using vector embedding techniques on handcrafted features~\cite{christlein2017writer} or hybrid approaches~\cite{christlein2015offline} until eventually deep learning dominated. 
Recent years have seen substantial advances through self-supervised representation learning, which reduces dependence on labeled training data~\cite{raven2024self}, and through efforts to make attributions interpretable rather than treating the classifier as a black box~\cite{raven2025interpretable}.
Tools designed with the paleographer as the intended user have begun to appear~\cite{grieggs2024paleographer}, reflecting growing recognition that deployment context shapes what a useful method looks like.
A persistent gap, however, separates laboratory performance from real archival deployment: published evaluations typically assume clean, exhaustively labeled datasets with a fixed set of scribes, while actual manuscript work involves degraded images, disputed or partial annotations, and open-set attribution problems.
Critically, existing systems offer their most confident output precisely where the answer is already largely known, whereas scholars most need computational assistance in cases of genuine uncertainty --- when a hand is unattested, when two attributions are plausibly equivalent, or when the material evidence is ambiguous.

\subsection{From writers to gender}
The work for identifying automatically the gender of writers has seen less attention than work looking into the particular identities of the hands.
In the communities of pattern recognition, handwriting analysis is a popular topic because it has many applications, but also because it exemplifies ideally a class of hard problems and is considered to a certain extent a benchmark for methods that might be more general in scope.
Yet the pattern recognition community is only too happy to externalize any questions as to what is gender, or even what is an individual.
From the point of view of pattern recognition, gender is a binary classification problem, writer identities are just labels in a dataset.
Curation of such datasets is where all nuance and ambiguity are suppressed.
Similarly, automatic methods researchers will very rarely venture into interpretation of their findings, of the weaknesses and strengths of the systems they develop in a way that is relatable to humanists.
While a few methods have addressed the problem of gender identification, none to our knowledge has provided any insights into why gender is distinguishable through writing.
Work presented by Yang et al.~\cite{yang2020men} in 2020 demonstrates through rigorous statistical analysis of fMRI scans at the time of writing that males and females activate significantly different parts of the brain when they write, as can be seen in figure~3(b) of their paper.
One might argue that we cannot extrapolate any conclusions from Chinese ideograms to western scripts.
These findings are compatible with arguments made by Marcelli in~\cite{marcelli2013some} about motor skills and development and how handwriting style contains many biometric features and is therefore largely immutable after a certain age.
Al Maadeed presented the KHATT database in~\cite{al2014automatic} and used metadata of the dataset to predict gender and ethnicity from handwriting samples.
While the results obtained are interesting, they refer only to Arabic handwriting, so only abstract conclusions from this study can be applied here.
In 2015 Djeddi et al. organized a blind competition including samples written in English and Arabic, one of whose two tasks was cross-lingual gender recognition~\cite{djeddi2015icdar2015}.
More recently, work was also introduced for Hebrew gender recognition~\cite{rabaev2022automatic}.
A comprehensive overview of publicly available datasets and a taxonomy of automatic methods has been provided by Rabaev and Litvak in~\cite{rabaev2023automated}.
One of the most important things to note is that the offline handwriting datasets that contain gender labels are all contemporary, and the larger ones are acquired from Middle Eastern and North African populations, typically university students.
Furthermore, all these datasets and studies are conducted under extremely controlled experimental conditions and would probably not be applicable to open-world real data, and even less so to historical manuscripts.
Yet identifying manuscripts written by female hands and documenting the acquisition of literacy by women in medieval times is an important question for historians.
\section{Corpus description}

\subsection{The Female Dominican monastery of SS.Pietro and Sebastiano in Naples}
\label{sec:monastery}
From here-on in this paper, for reasons of brevity, we will refer to this corpus as the Naples Corpus.
The corpus originates from the Female Dominican monastery of SS.\,Pietro e Sebastiano (Saints Peter and Sebastian) of Naples.
The convent was originally called San Pietro a Castello and was founded in the fourteenth century by Mar\'ia of Hungary and Charles~II d'Anjou as a residence for noblewomen close to the Angevin court, situated near Castelnuovo, the royal residence~\cite{ambrosio2015literacy}.
After the area around Castelnuovo was destroyed during the conflicts of the early fifteenth century, the community relocated to the older Benedictine monastery of San Sebastiano, taking the double name it still bears today.
In the second half of the fifteenth century the convent adopted the Observance, a strict reform movement within the Dominican Order, which had direct consequences for how the community managed its documentation~\cite{ambrosio2024turning}.

Before the 1470s, the compilation of the convent's account books and administrative registers was dominated by friars and procurators from the nearby monastery of San Domenico Maggiore, to whom the \emph{cura monialium} --- formal pastoral and administrative care of the nuns --- had been entrusted~\cite{ambrosio2015literacy,ambrosio2024turning}.
These men brought with them well-developed scribal traditions, writing in notarial minuscule, \emph{mercantesca}, and humanist scripts~\cite{ambrosio2024turning}.
In 1466, Marziale Auribelli, Master General of the Order, issued \emph{Ordinationes} ordering the prioress to designate ``duas vel tres de sororibus ad hoc'' --- two or three nuns --- to take charge of the convent's written records, a measure prompted by damage to the conventual patrimony caused by imprudent friar management~\cite{ambrosio2015literacy,ambrosio2024turning}.
From 1471 onwards, nuns participated in record-keeping with increasing regularity, marking a documented transition to autonomous administration~\cite{ambrosio2024turning}.
The nuns wrote with varying and generally more modest levels of scribal skill than their friar predecessors, ranging from a practised cursive to elementary, hesitant script~\cite{ambrosio2015literacy}.

Nine account registers from this convent survive at the Archivio di Stato di Napoli (ASN, Corporazioni religiose soppresse, 1395--1402) --- an exceptional number for any Neapolitan women's convent~\cite{ambrosio2015literacy}.
We annotated three of these registers, chosen to span the period of institutional transition described above; their key properties are summarised in Table~\ref{tab:registers}.
ASN~436 (1473--1478) is written predominantly by friars and procurators of San Domenico Maggiore; it has few identifiable hands and no gender annotations.
ASN~1397 covers the transitional period from the 1470s through the 1490s (87 pages) and contains mixed nun and friar hands; gender labels are not available with sufficient certainty for quantitative use.
ASN~1401 records daily kitchen and pantry expenses from 1485 to 1496 across 385 pages; it has 8 identified hands (5 female, 3 male) with reliable gender annotations, making it the only register suitable for the classification experiments reported in this paper.
The male hands in ASN~1401 belong to friars or procurators of San Domenico Maggiore who retained a residual oversight function; these men remained outside the \emph{rota}, the rotating wheel set in the convent wall through which objects and documents were exchanged with the outside world~\cite{ambrosio2015literacy}.
We make this institutional history explicit because, as we discuss in \S\ref{sec:challenges}, it produces a spatial confound with direct consequences for the interpretation of automatic gender classification.

\begin{table}[t]
  \centering
  \caption{The three annotated registers of the Naples Corpus.
           Only ASN~1401 is used for the classification experiments reported in this paper.}
  \label{tab:registers}
  \begin{tabular}{lrrlll}
    \toprule
    \textbf{Register} & \textbf{Pages} & \textbf{Hands} & \textbf{Period} & \textbf{Primary scribes} & \textbf{Gender labels} \\
    \midrule
    ASN 436  &  --- & few  & 1473--1478 & Friars (San Domenico Maggiore) & No  \\
    ASN 1397 &  87  & several & 1470s--1490s & Mixed (nuns + friars)          & No  \\
    ASN 1401 & 385  & 8    & 1485--1496 & Nuns (5F, 3M oversight)        & Yes \\
    \bottomrule
  \end{tabular}
\end{table}

\subsection{Data annotation}
The corpus was annotated so that quantifiable measurements about any kind of observation and analysis can be systematized.
Data annotation was guided by two pragmatic constraints: minimizing annotation effort and reusing existing tools for annotation without data preprocessing or curation before annotation.
These constraints led us to the decision of modeling the annotation scheme as simply defining rectangular regions over each page image such that they are tightest possible around a text region.

\begin{figure}[ht!]
    \centering
\begin{tabular}{cc}
    \includegraphics[width=0.45\linewidth]{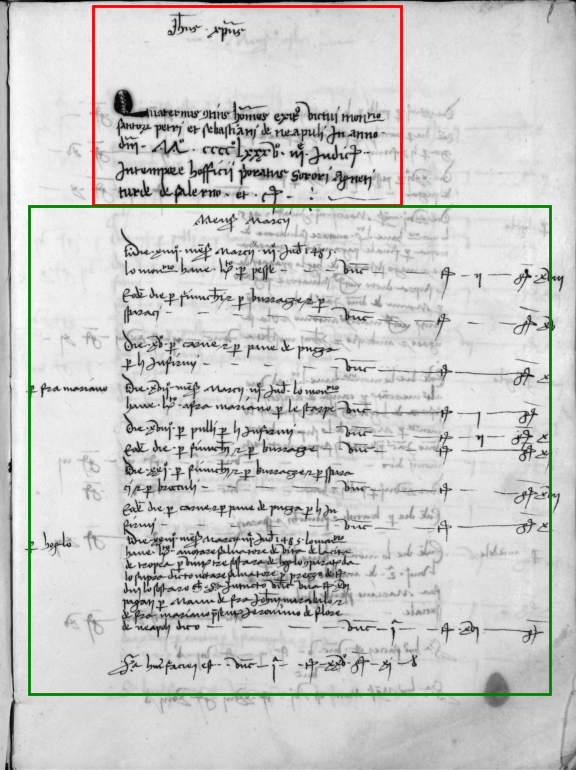} &
    \includegraphics[width=0.45\linewidth]{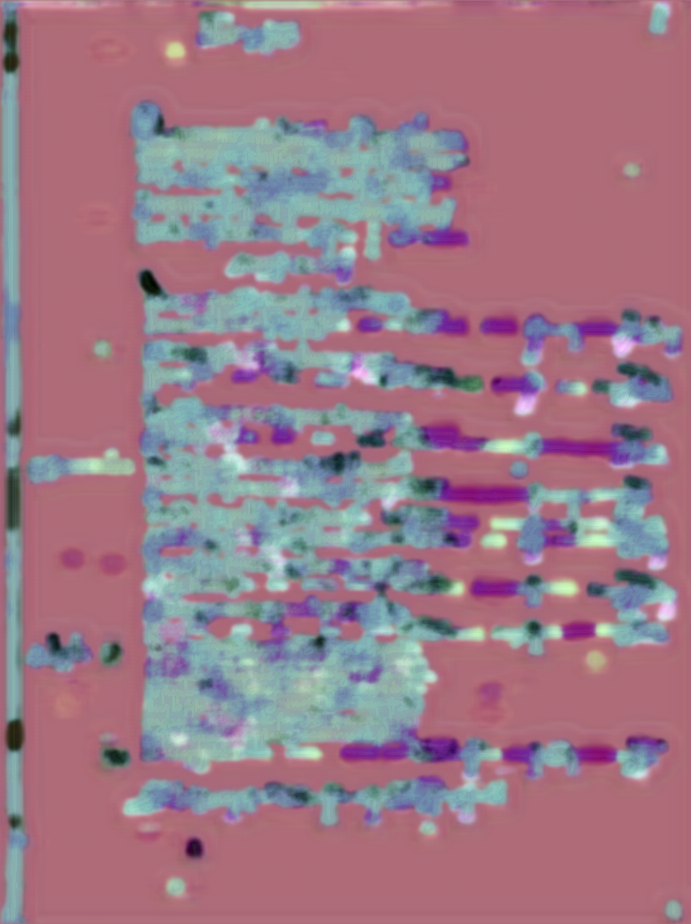} \\
\end{tabular}
    \caption{Indicative annotation of two hands on one page and a qualitative texture map}
    \label{fig:1401_example}
\end{figure}

\begin{figure}[hb!]
    \centering
    \begin{tabular}{ccccc|cc}
    \includegraphics[trim={148px 37px 135px 44px}, clip,width=.135\textwidth]{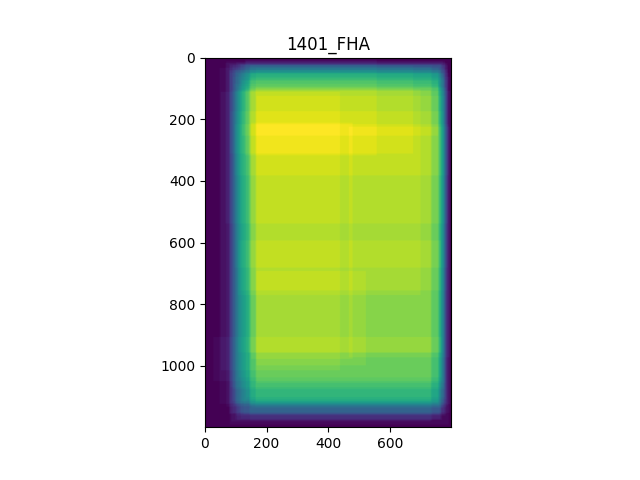} &
    \includegraphics[trim={148px 37px 135px 44px}, clip,width=.135\textwidth]{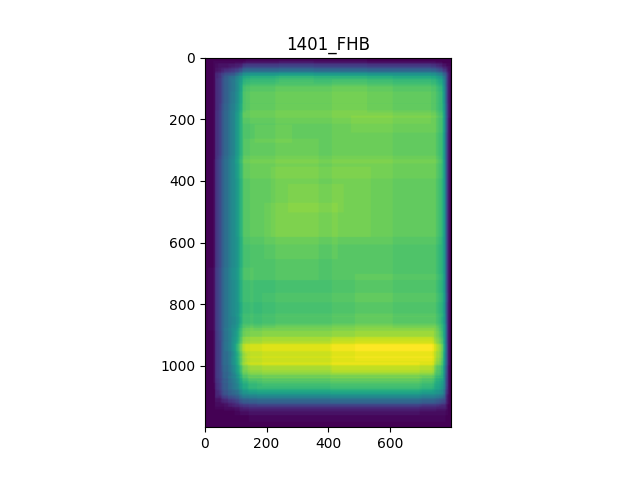} &
    \includegraphics[trim={148px 37px 135px 44px}, clip,width=.135\textwidth]{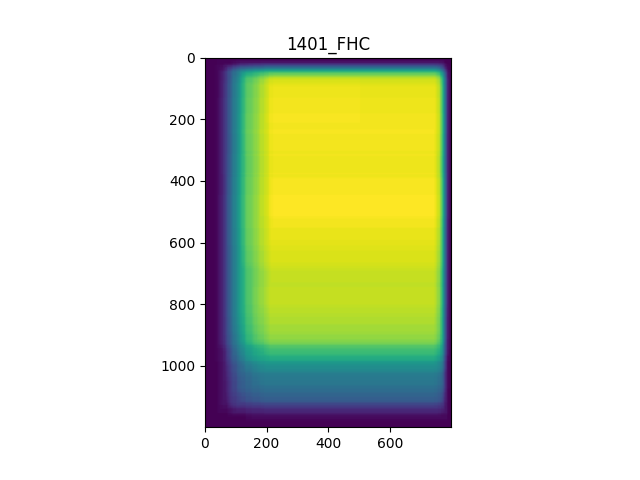} &
    \includegraphics[trim={148px 37px 135px 44px}, clip,width=.135\textwidth]{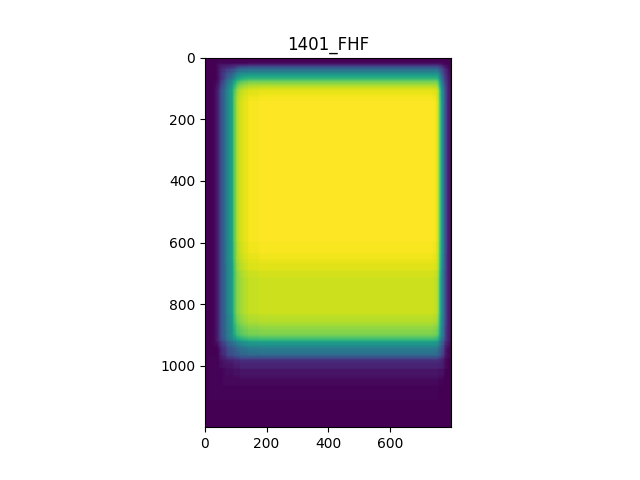} &
    \includegraphics[trim={148px 37px 135px 44px}, clip,width=.135\textwidth]{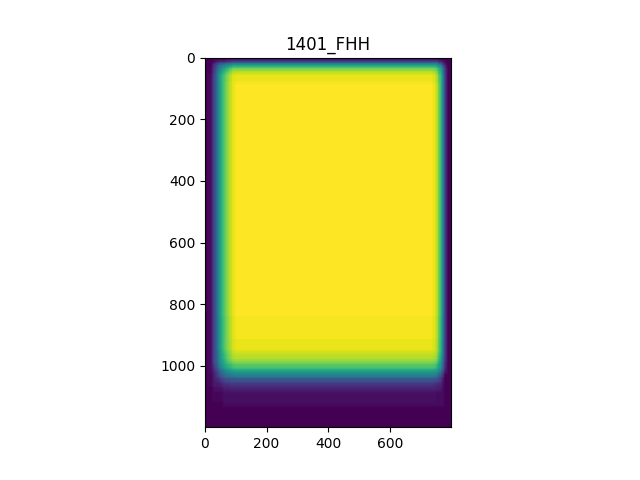} &
    \includegraphics[trim={148px 37px 135px 44px}, clip,width=.135\textwidth]{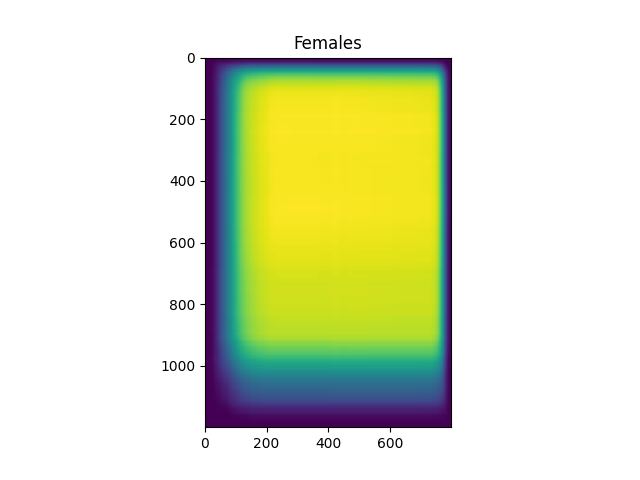} &
    \includegraphics[trim={148px 37px 135px 44px}, clip,width=.135\textwidth]{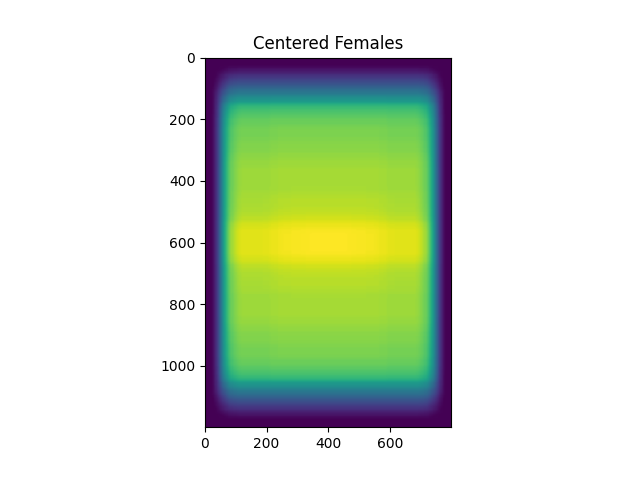} \\
    Female A & Female B & Female C & Female F & Female H & All & Centered \\

    \includegraphics[trim={148px 37px 135px 44px}, clip,width=.135\textwidth]{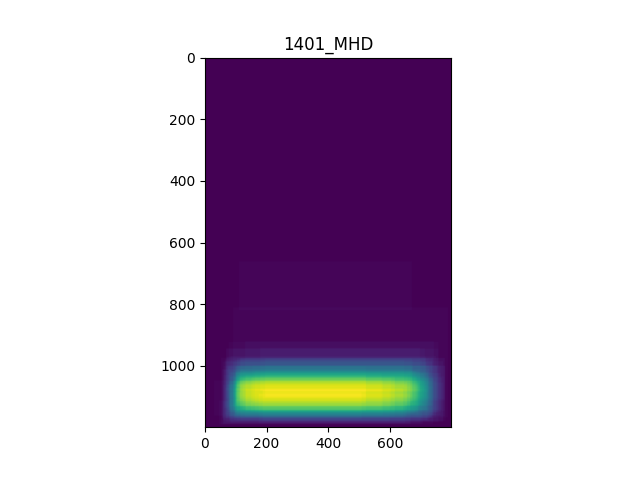} &
    \includegraphics[trim={148px 37px 135px 44px}, clip,width=.135\textwidth]{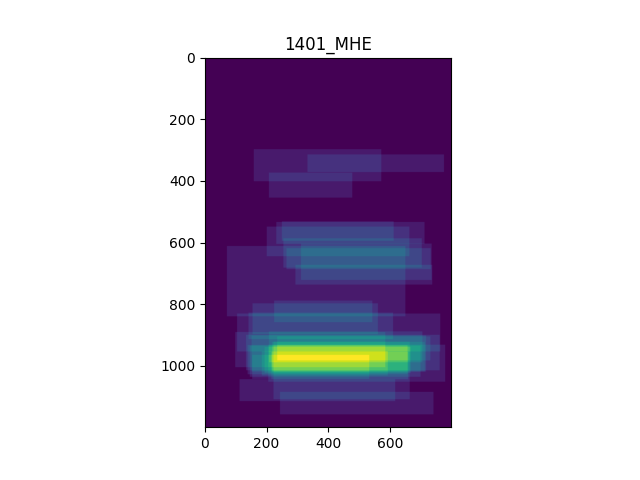} &
    \includegraphics[trim={148px 37px 135px 44px}, clip,width=.135\textwidth]{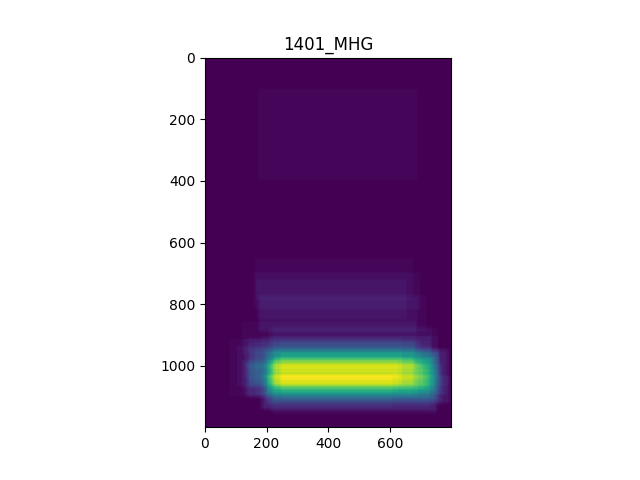} &
    &
    &
    \includegraphics[trim={148px 37px 135px 44px}, clip,width=.135\textwidth]{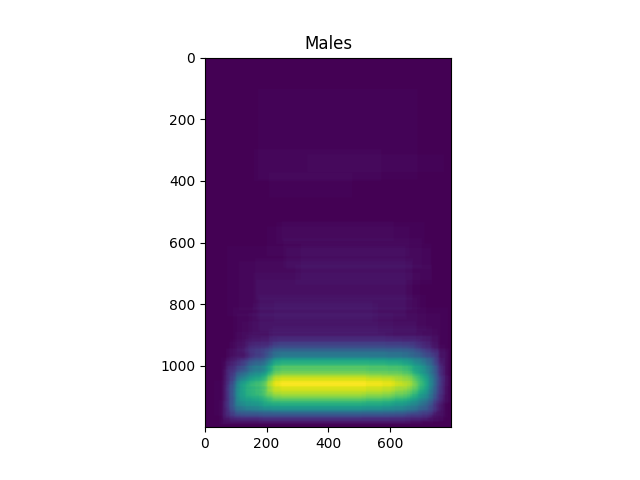} &
    \includegraphics[trim={148px 37px 135px 44px}, clip,width=.135\textwidth]{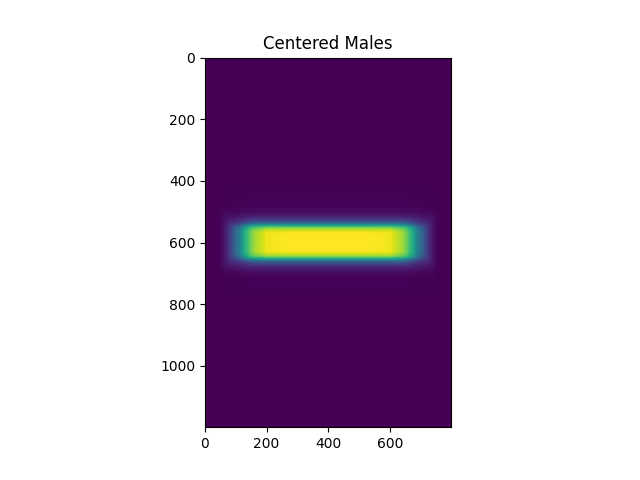} \\
    Male D & Male E & Male G & & & All & Centered
    \end{tabular}
    \caption{%
        Spatial footprint of every annotated hand in the corpus.
        Female hands (top row) are the nuns who compiled the register and occupy the main body of each page.
        Male hands (bottom row) belong to friars or procurators of San Domenico Maggiore who retained a supervisory role and appear as narrow bands at the foot of pages.
        This structural asymmetry is an institutional artefact rather than a reflection of handwriting-style differences between male and female scribes.}
    \label{fig:genderroles}
\end{figure}

\subsection{Corpus challenges and visual characteristics}
\label{sec:challenges}
While the data has been annotated with hand id and its estimated gender, it might be too simplistic to perceive this data as hand and gender identification.
As can be seen in Fig.~\ref{fig:1401_example} the paper is quite thin and bleed-through is a major challenge specifically for binarization.
Furthermore many pages have cross-out and corrected text which adds a lot of noise to writer identification methods.
Another significant problem is discriminating text from horizontal lines marking table geometry, as can be seen in Fig.~\ref{fig:1401_example} these horizontal lines appear as the most prominent texture differentiation.
This data highlights how the difference between what is text-content and what is text-style becomes harder to distinguish as the text sampling becomes more granular.

A further structural confound follows directly from the institutional history described in \S\ref{sec:monastery}.
Because the hands labelled as male appear almost exclusively as brief entries at the foot of pages --- consistent with the residual oversight role of the friars from San Domenico Maggiore --- while female hands occupy the main body of text, any classifier with access to spatial position will tend to separate the labelled genders by page location rather than by handwriting style.
This spatial asymmetry is clearly visible in Fig.~\ref{fig:genderroles}.
We therefore caution that results reported as \emph{gender} identification in this corpus may more precisely reflect identification of \emph{scribal role} as conditioned by the specific institutional history of this monastery.
Whether scribal role and gender are separable in this context is a question we leave to historians of the period.
\section{Method}
\subsection{Constraints and motivation}
Our main motivation is to create tools that assist paleographers rather than trying to imitate them, let alone replace them.
Paleographers want assistance with data they are uncertain about, meaning that in the use cases they care about there is no such thing as ground truth, at least not in the absolute sense that permits clean experimental evaluation.
The most common workaround is self-supervision under the assumption of one scribe per page or per document.
The Naples corpus cannot be served that way because it has many hands per document and per page; the quantum of unique hand-ness is extremely low, often at the level of a single text line.
Motivated by these constraints, we focused on knowledge-transfer strategies, which allow the potential biases of the training data to be externalized to a synthetic domain, making it harder for them to interfere with paleographic inference.
The fine granularity of the analysis also makes fully convolutional architectures almost mandatory, since dense patch sampling would be prohibitive and would discard spatial context.
To the best of the authors' knowledge there is no established fully convolutional method for dense texture analysis of document images.
Our strategy was therefore to train a network on synthetic data with the inference characteristics we require.
We chose to implement a Multi-Task Learning (MTL) fully convolutional model because MTL is known to improve generalization to new tasks by forcing shared representations to support multiple objectives simultaneously~\cite{caruana1997multitask}.

\subsection{Synthetic Data}
Synthetic data has been proven sufficient to build powerful word-image classification neural networks since 2014~\cite{jaderberg2014synthetic}.
In 2016 Gupta et al. employed synthetic data to create large-scale supervision for scene-text segmentation~\cite{gupta2016synthetic}.
In our approach we separated the synthesis problem into two parts: first rendering relatively simple pseudo-pages with their respective segmentation labels, and then applying Tormentor, a rich set of text-specific augmentations that are automatically applicable to segmentation masks~\cite{nicolaou2022tormentor}.

\begin{figure}
    \centering
    \begin{tabular}{cc}
    \includegraphics[width=.45\textwidth]{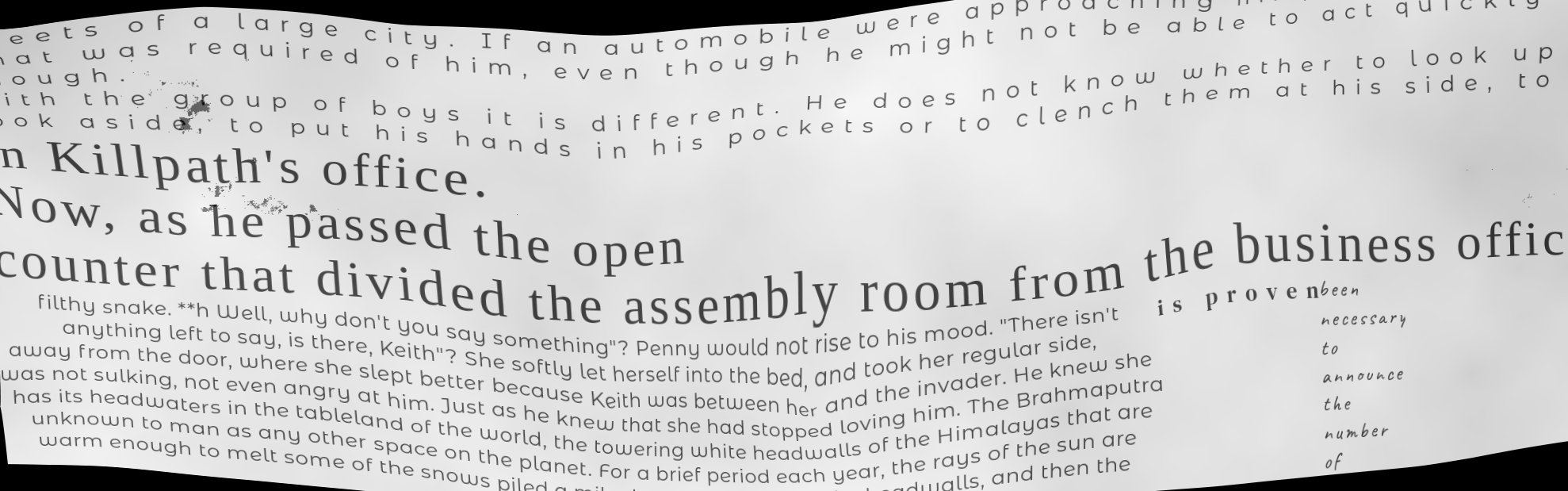} &
    \includegraphics[width=.45\textwidth]{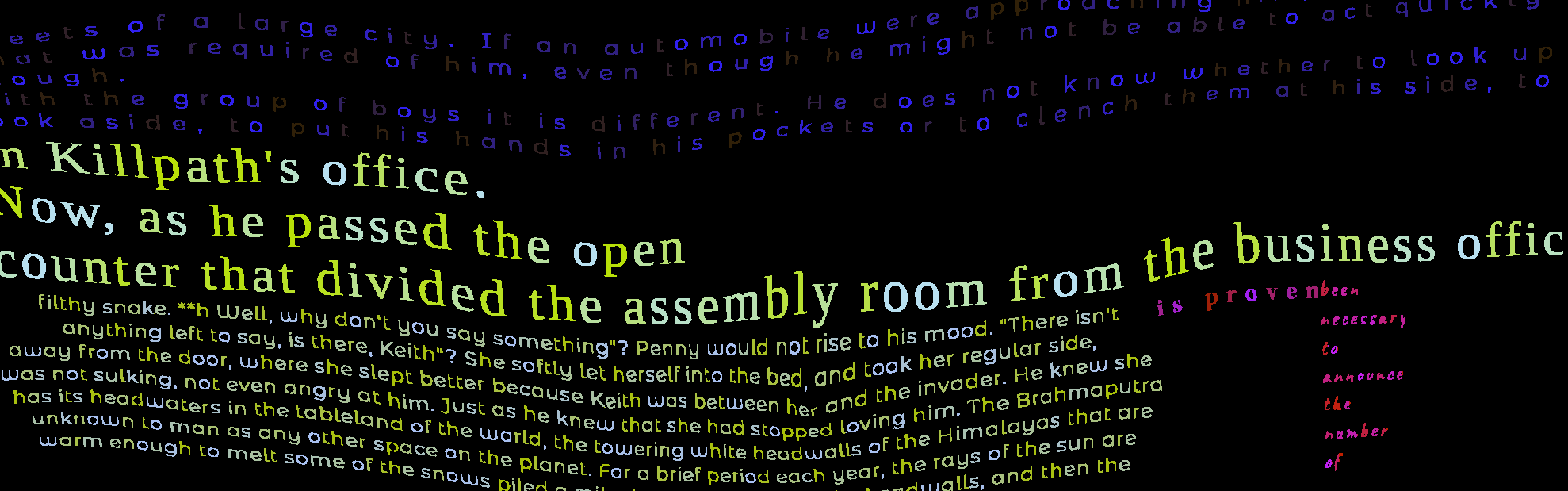} \\
    \includegraphics[width=.45\textwidth]{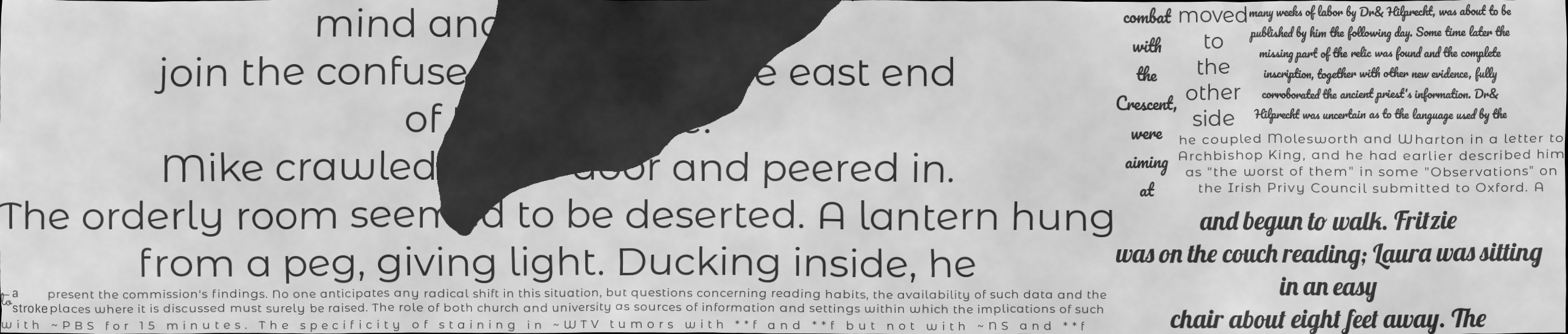} &
    \includegraphics[width=.45\textwidth]{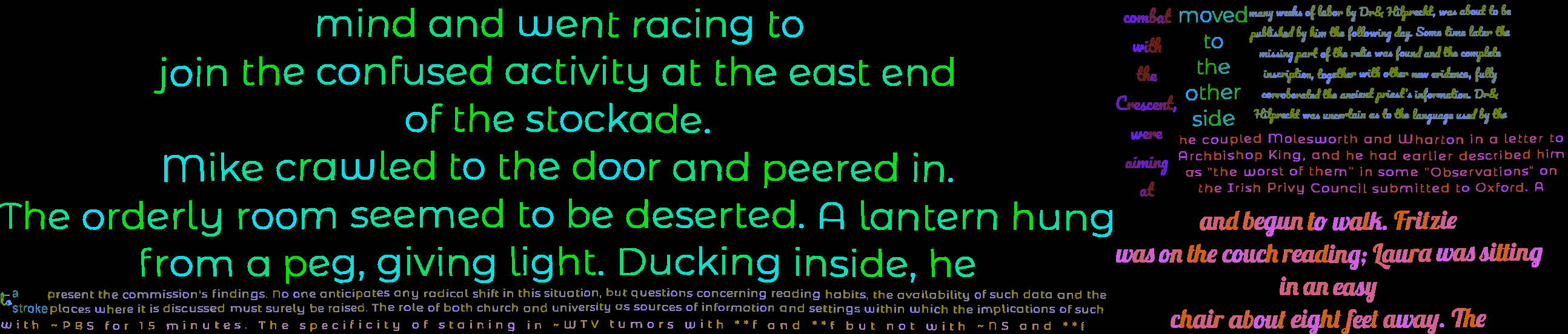} \\
    \includegraphics[width=.45\textwidth]{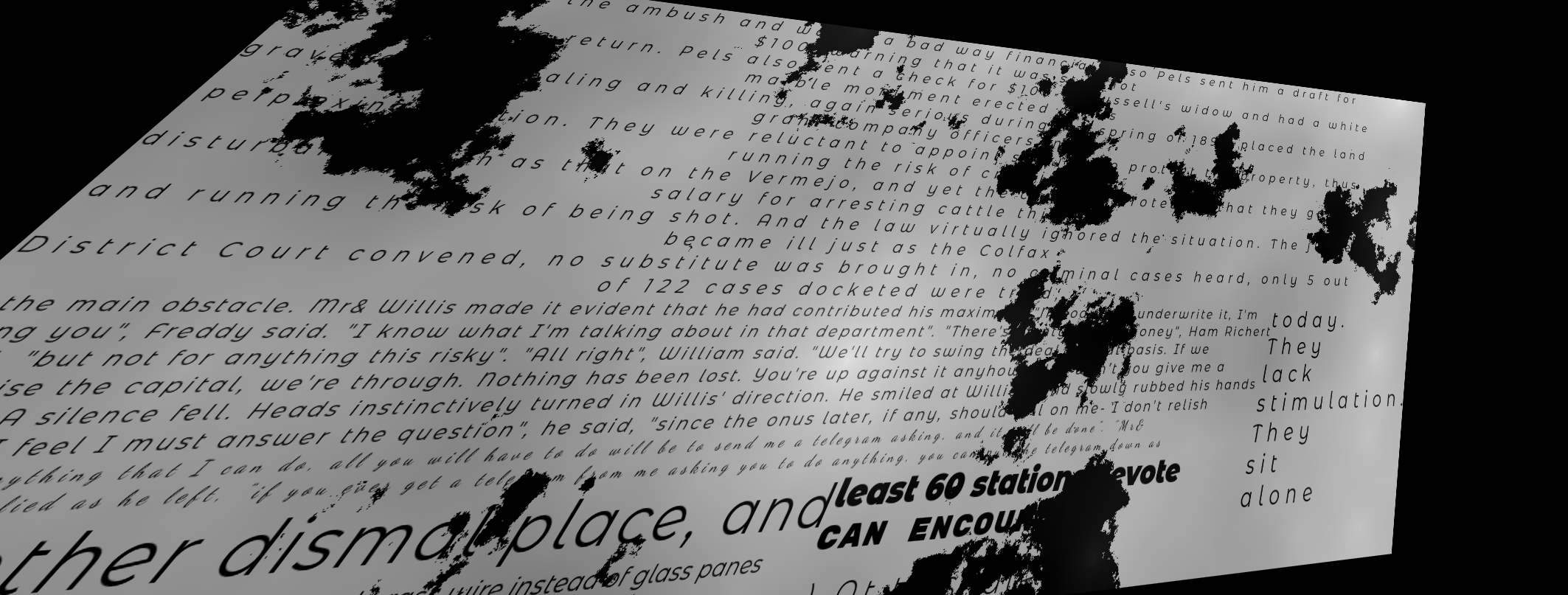} &
    \includegraphics[width=.45\textwidth]{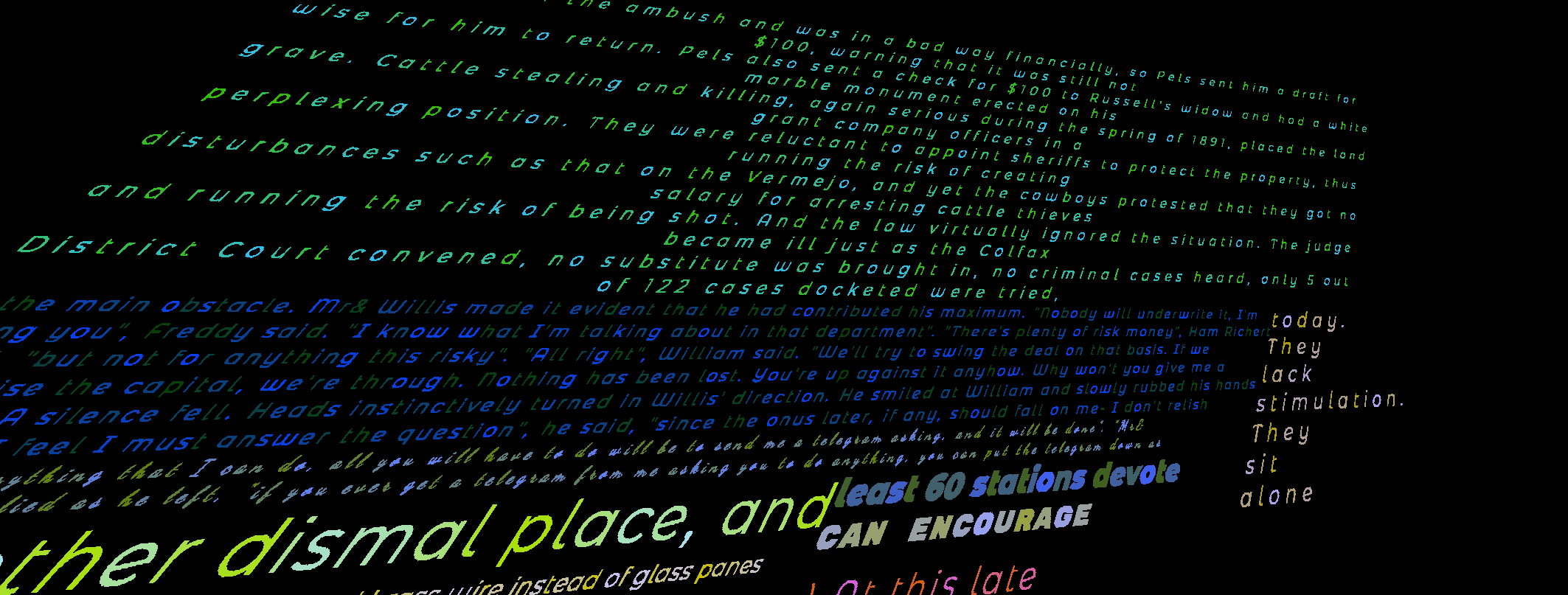}
    \end{tabular}
    \caption{Synthetic pseudo-pages (left) and their multiplexed segmentation maps (right).
    The segmentation maps encode font family, font size, and character identity in the RGB channels.}
    \label{fig:synthdata}
\end{figure}

The pseudo-pages were designed to incorporate text-style variations within a single image while preserving the rectangular, line-based notion of text.
A pseudo-page has between 4 and 10 pseudo-paragraphs.
A pseudo-paragraph consists of a rectangular block of text in a single text style.
We employed the Brown Corpus~\cite{francis1979brown} to sample text blocks in order to have realistic $n$-grams in the data.
The area of each pseudo-paragraph is first determined at random; then its text style is set by independently sampling a font family, font size, kerning, and horizontal alignment from custom distributions whose design is the critical step in text-image synthesis~\cite{jaderberg2014synthetic}.
Text was rendered in a single operation with word-wrapping on the paragraph width using Pango with PyCairo.
The authors found these tools to be superior for this use-case to all free popular alternatives as they provide both high-quality atomic rendering and character-level bounding boxes.
Routing and chaining of elementary fractal augmentations, both ventral and dorsal, are used to simulate complex degradations such as stains, reflections, perspective distortion, wrapping, shadows, and camera blur.
In Fig.~\ref{fig:synthdata} samples from the final synthetic dataset together with their ground-truth maps can be seen.
The actual classes we deemed most informative were font family, font size, and character; they have been multiplexed into the RGB color space.
As can be seen, letter colors are consistent within a pseudo-paragraph but differ across paragraphs, providing a strong per-pixel supervision signal for the metric-learning loss described below.
In the present setup, pseudo-pages are between half and two megapixels.

\subsection{TextileNet}
\begin{figure}
    \centering
    \includegraphics[width=.8\linewidth]{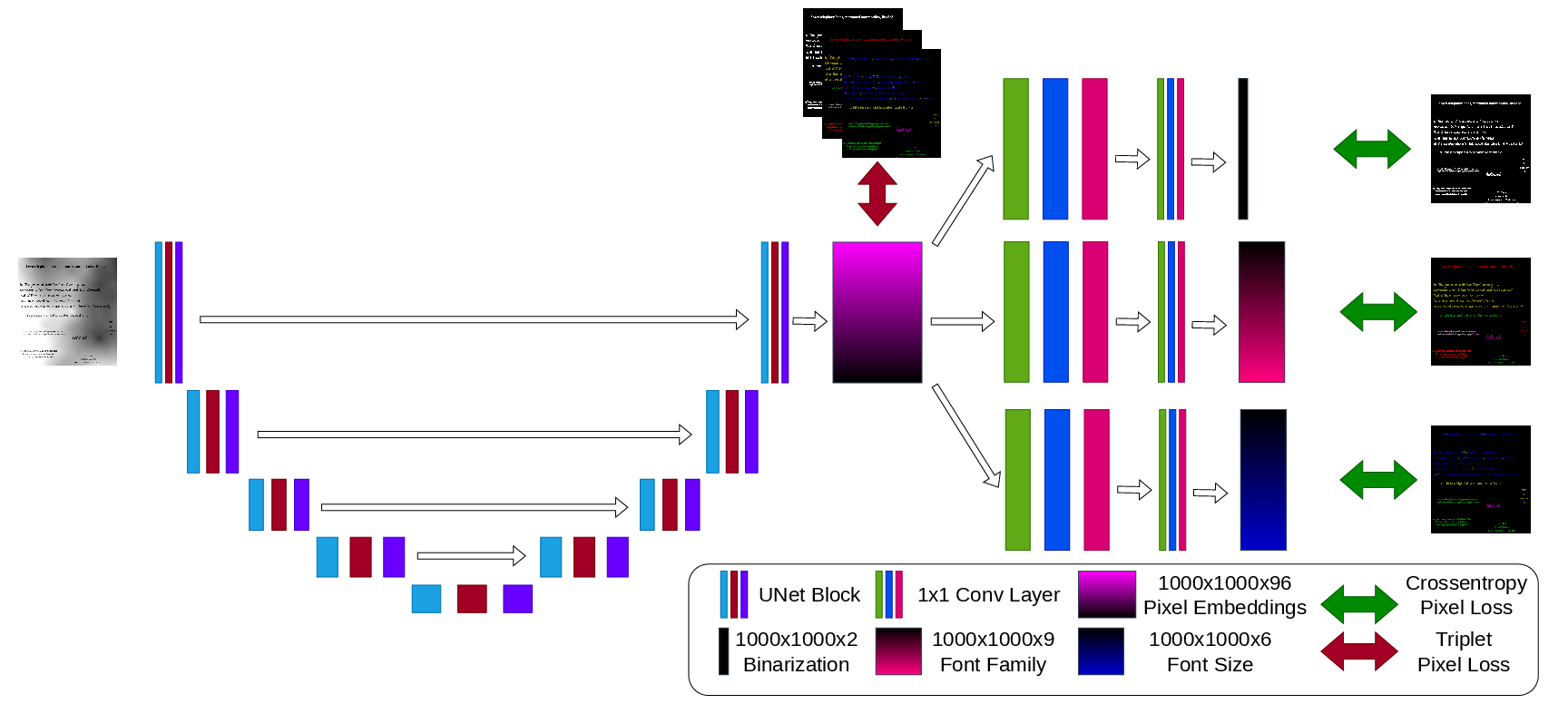}
    \caption{TextileNet Architecture.}
    \label{fig:textile}
\end{figure}

The proposed TextileNet is fully convolutional, intended to run directly on whole pages, and consists of an IUnet~\cite{etmann2020iunets} backbone for memory efficiency, which produces pixel vectors of 384 dimensions.
All classification heads are implemented as a cascade of $1\times1$ convolutions acting as multilayer perceptrons over the pixel embeddings produced by the IUnet, so that the IUnet is the only component that can learn spatial geometry.
A foreground/background head was added to make the model self-contained, since binarization is useful for several downstream inference schemes and also assists the other heads in foreground-background segmentation.
A font-family head and a font-size head are always present; a character head is implemented in some variants but requires GPUs with more than 40~GB of memory for training.
Training is performed with a batch size of 1 and all heads are trained simultaneously on each sample using a cross-entropy loss on the segmentation masks.
In Fig.~\ref{fig:textile} the architecture can be seen.

\subsection{Pixel-level Relative Ratio Triplet Loss}
Initially, training TextileNet converged to large validation errors on the segmentation heads other than foreground/background; the orthogonality of the font-family and font-size tasks appeared to cause competing gradients to cancel each other out, and the IUnet received no meaningful learning signal from the segmentation heads.
We introduced a pixel-level metric loss partially based on the ratio loss~\cite{balntas2016learning}, treating the pixel embeddings of the entire image as a dataset on which metric learning is performed.
We select all foreground pixels up to a maximum of 50,000, sampled at random.
Two pixels are considered to share the same identity if they agree on all segmentation labels.
We compute a pairwise distance matrix $D \in \mathbb{R}^{P \times P}$ over the texture embedding layer activations, and use it to construct a hard-positive linkage matrix $A\in \{0,1\}^{P \times P}$ and a hard-negative linkage matrix $H\in \{0,1\}^{P \times P}$, having ones at the hardest positive and hardest negative sample for each anchor pixel.
Hard positives and negatives are computed via argsort and require no gradient.
The loss is defined in (\ref{eq:metricloss}); introduction of this pixel-level triplet loss made training the character head possible and reduced validation errors on the other tasks by more than a factor of five as can be seen in table~\ref{tab:textilenet_val}.
\begin{equation}
\mathcal{L}
=
\frac{\mathrm{scale}}{P}
\sum_{i=1}^{P}
\left(
\frac{
\sum_{j=1}^{P} D_{ij} A_{ij}
}{
\sum_{j=1}^{P} D_{ij} A_{ij}
+
\sum_{j=1}^{P} D_{ij} H_{ij}
+
\varepsilon
}
\right)^2
\label{eq:metricloss}
\end{equation}
\section{Experiments}

\subsection{TextileNet Validation}
TextileNet performance on synthetic data is not directly applicable to paleography, but its performance on held-out unseen data helps build credibility for knowledge transfer and zero-shot employment on manuscripts.
Table~\ref{tab:textilenet_val} reports pixel error rates for three variants differing in embedding dimension, use of the pixel-level triplet loss, and validation augmentation regime.
The most important observation is the effect of the triplet loss: without it, font-family and font-size error rates of $8.40\%$ and $7.17\%$ indicate that the IUnet backbone receives no useful gradient from those heads, while adding the loss reduces them to $0.32\%$ and $0.47\%$ respectively.
The full TextileNet model shows slightly higher errors on foreground, font-family, and font-size compared with the 192-d triplet variant; this is expected both because the final model additionally supervises the character head, which is a substantially harder problem, and because the routing augmentation used at validation represents a harder distribution than the flat augmentation used for the smaller variants.

\begin{table}[!ht]
  \centering
  \caption{Pixel error rates (\%) of TextileNet variants on held-out synthetic validation sets.
    \emph{Flat}: no routing augmentation at validation (matching the smaller training regime).
    \emph{Routing}: full routing-and-chaining augmentation used for the final TextileNet.
    A horizontal rule separates variants with different model capacity and validation conditions.
    Best result per column in \textbf{bold}.}
  \label{tab:textilenet_val}
  \begin{tabular}{cc@{\hspace{0.5em}}l@{\hspace{2em}}rrrr}
    \toprule
    \multicolumn{3}{c}{\textbf{Variant}} & \multicolumn{4}{c}{\textbf{Performance (\% error)}} \\
    \cmidrule(r){1-3} \cmidrule(l){4-7}
    \textbf{Emb.~dim} & \textbf{Triplet} & \textbf{Augment.} & \textbf{FG} & \textbf{F.~Family} & \textbf{F.~Size} & \textbf{Char} \\
    \midrule
    192 & $\times$      & Flat    & $\mathbf{0.08}$ & $8.40$ & $7.17$ & --- \\
    192 & $\checkmark$  & Flat    & $0.37$ & $\mathbf{0.32}$ & $\mathbf{0.47}$ & --- \\
    \midrule
    384 & $\checkmark$  & Routing & $0.60$ & $1.51$ & $0.90$ & $\mathbf{0.94}$ \\
    \bottomrule
  \end{tabular}
\end{table}

\begin{figure}
    \centering
    \begin{tabular}{cccc}
    \includegraphics[width=.23\textwidth]{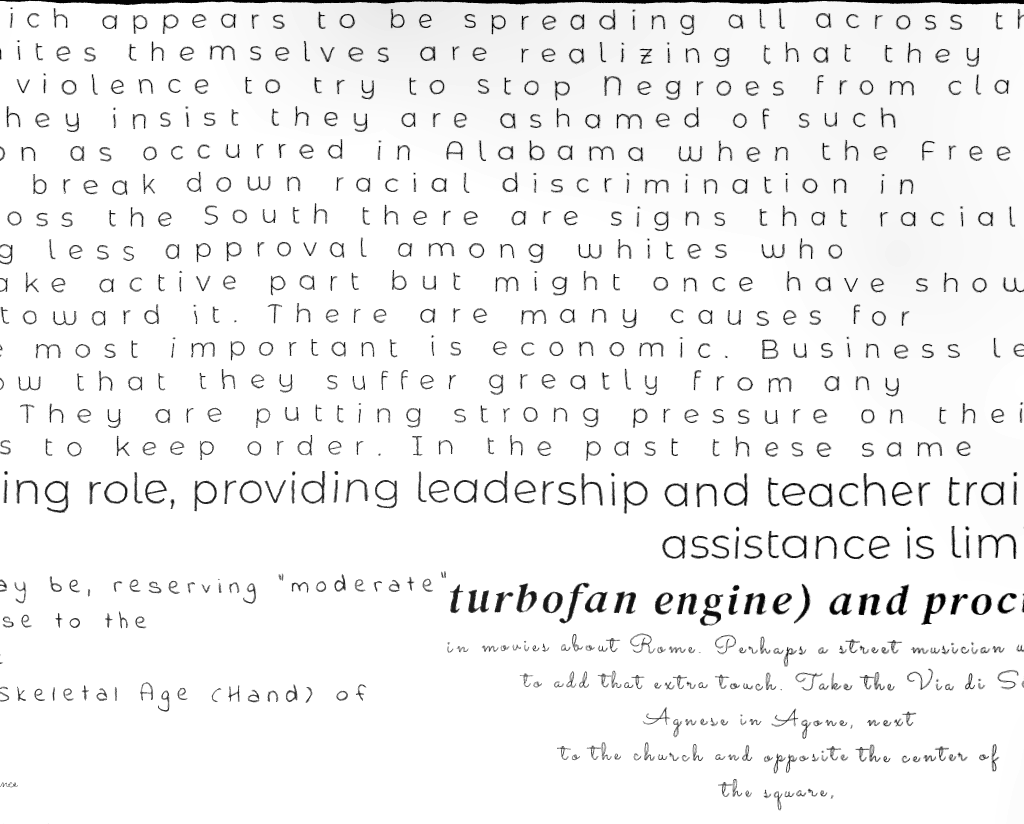} &
    \includegraphics[width=.23\textwidth]{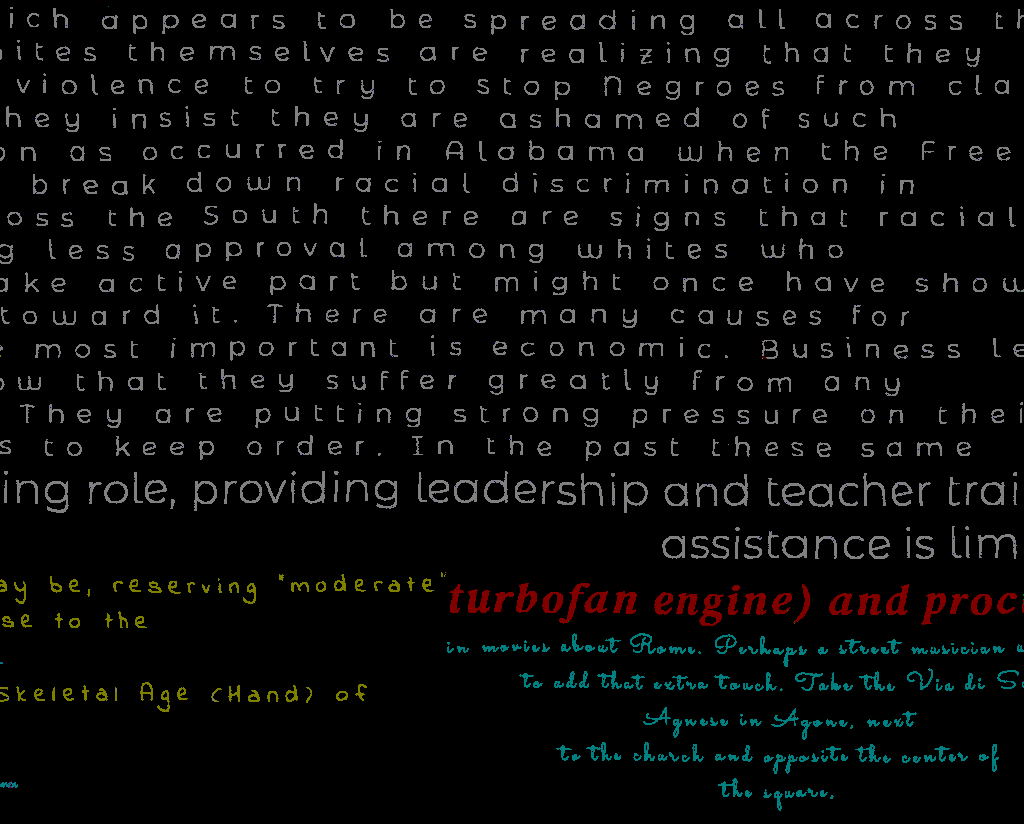} &
    \includegraphics[width=.23\textwidth]{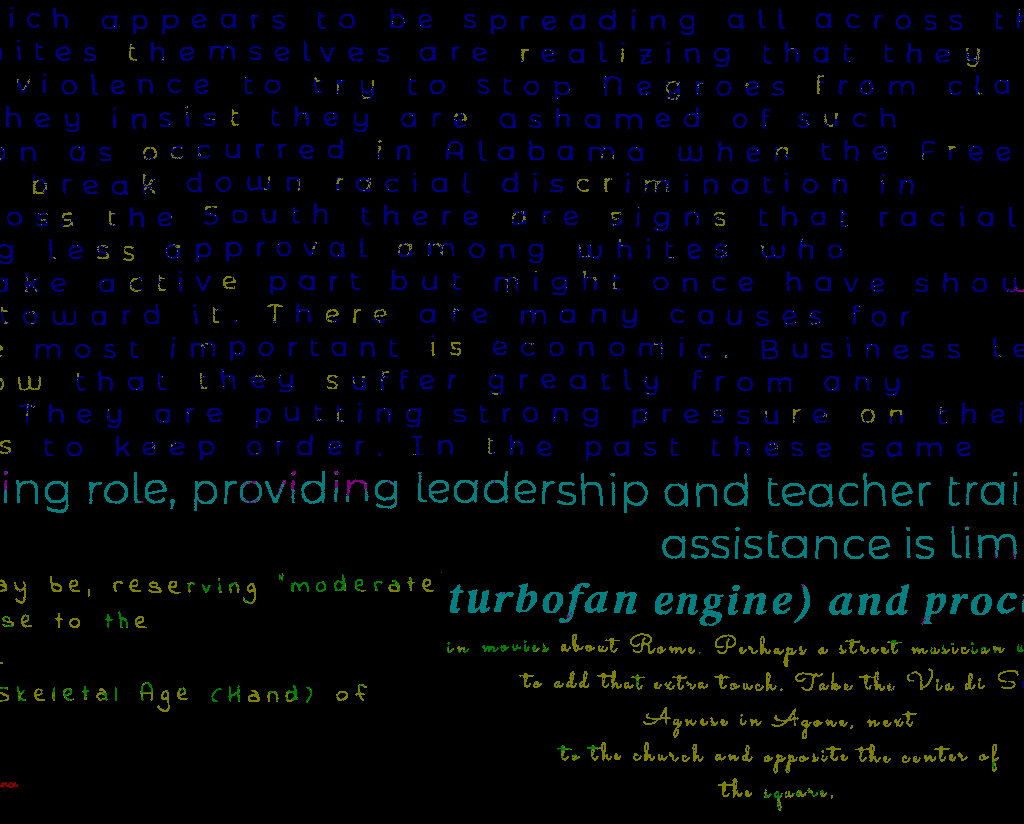} &
    \includegraphics[width=.23\textwidth]{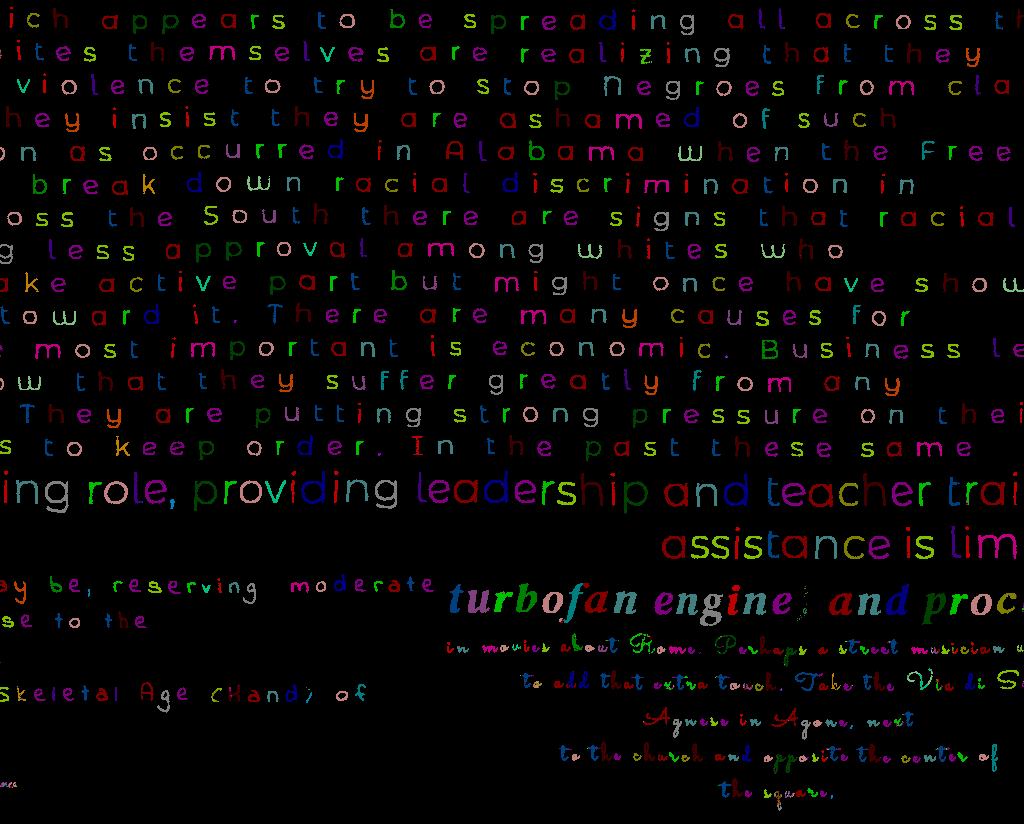} \\
    Input Image & F. Family & F. Size & Char \\
    \includegraphics[width=.23\textwidth]{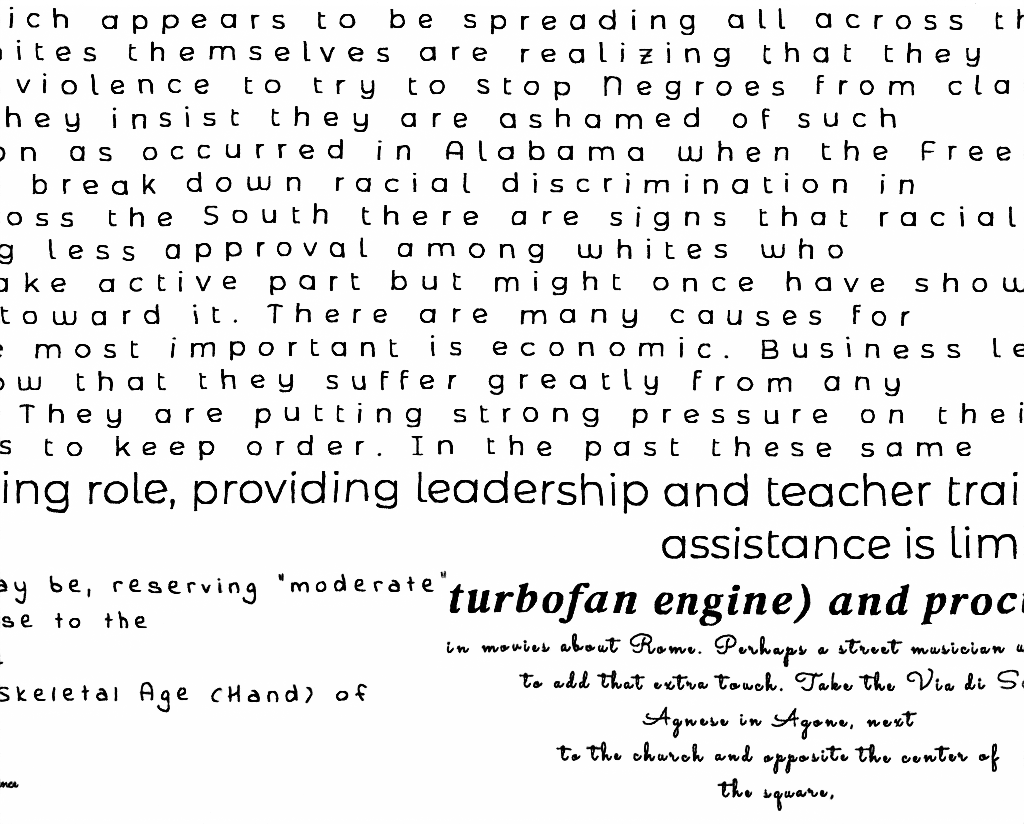} &
    \includegraphics[width=.23\textwidth]{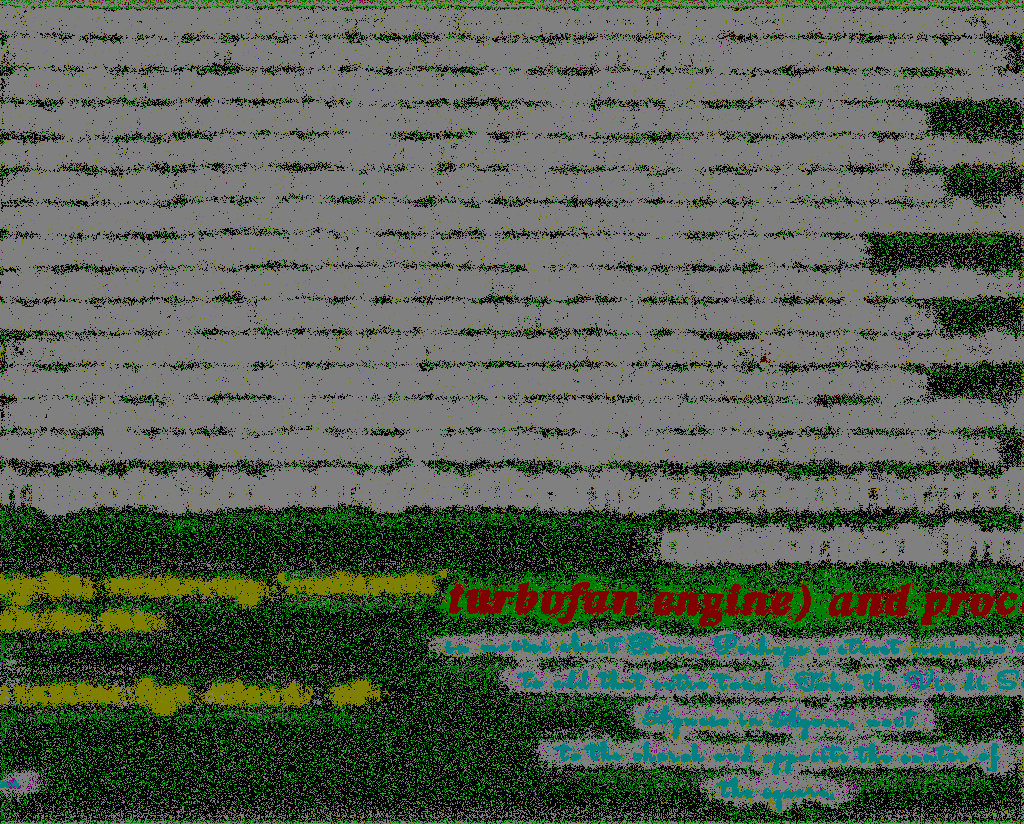} &
    \includegraphics[width=.23\textwidth]{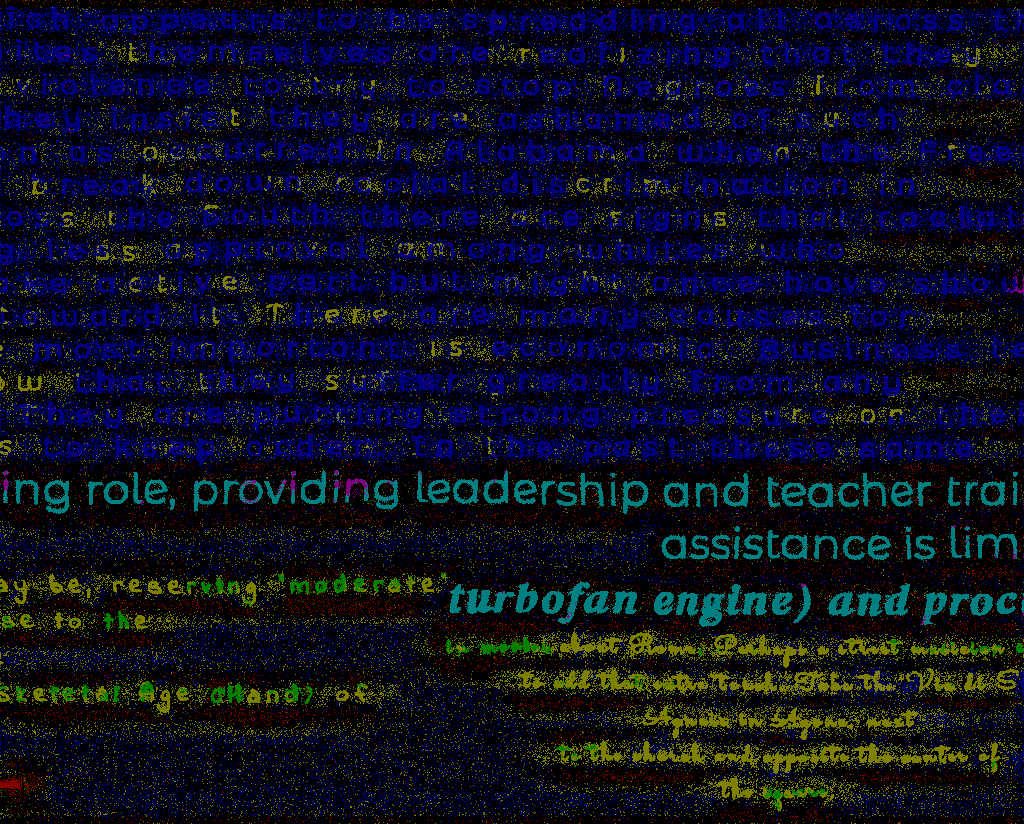} &
    \includegraphics[width=.23\textwidth]{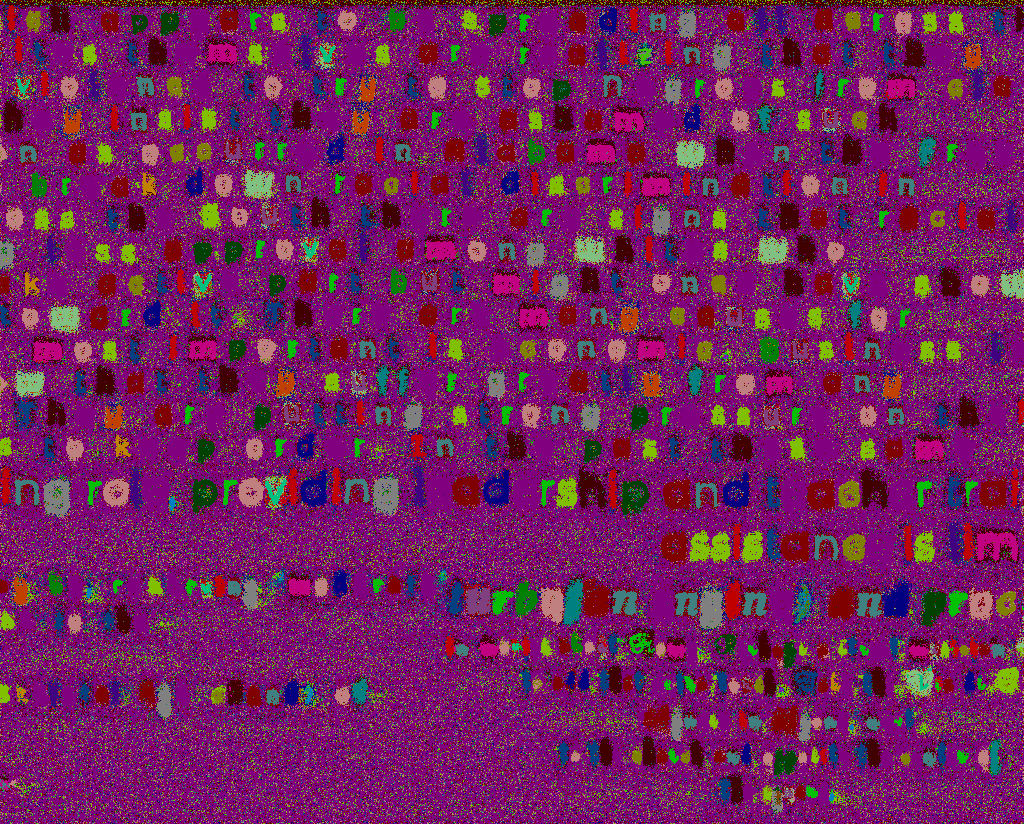} \\
    Binarization & F. Family No BG & F. Size No BG & Char No BG \\
    
    \end{tabular}
    \caption{Indicative TextileNet outputs for all heads on a synthetic validation sample.
    No-BG maps suppress the background logit before softmax, suggesting an effective receptive field of roughly 40~pixels radius.}
    \label{fig:textilenetvalidation}
\end{figure}

\subsection{Paleographic User Study}
As we were unaware of established baselines or even established evaluation protocols, we designed a user experiment in order to see how humans would perform in this limited visual and diplomatic context.\footnote{The quiz is still open and accessible at \url{https://docs.google.com/forms/d/1-8cPZ986z24HNe1Iz5sjjDREkH2vL-tpb1P56JWNDHs}}
Designing the experiment was influenced by many needs often conflicting: making it agreeable and rewarding for the participants, making a test a machine could also take, making it unambiguous, and economizing the participants' time.
Two kinds of questions were devised: a simple question where given a pair of samples participants have to reply if the samples come from the same scribe or not; and a triplet question where given an anchor sample and a pair of samples the participant must answer if the first or the second sample matches the anchor.
The questions and the modalities were balanced so that they have a $50\%$ chance of being correct by random choice.
Each sample comes from a unique page so that no two samples in the same question come from the same page.
After pairs and triplets were randomly selected, the 80 questions retained were chosen under the condition that all samples contain at least one three-letter word.
The users were instructed to be as fast as possible, indicatively 5--10 seconds per question, but it is our understanding that most participants took more time.

\begin{figure}[hbt!]
\centering
\begin{tabular}{ccc}
     \includegraphics[width=.26\textwidth]{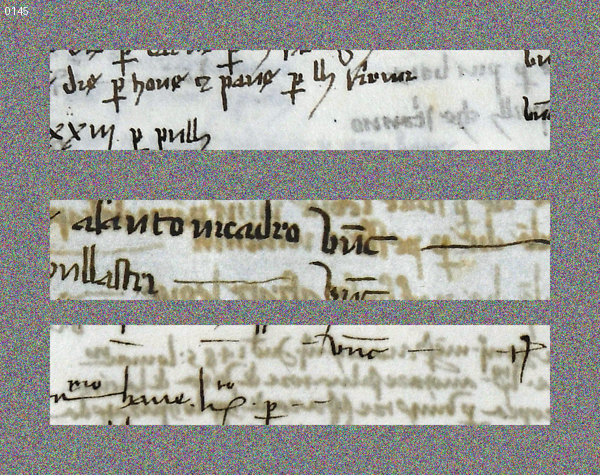} &
     \includegraphics[width=.35\textwidth]{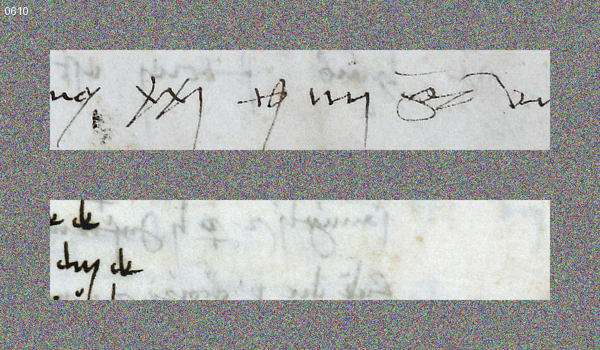} &
     \includegraphics[width=.26\textwidth]{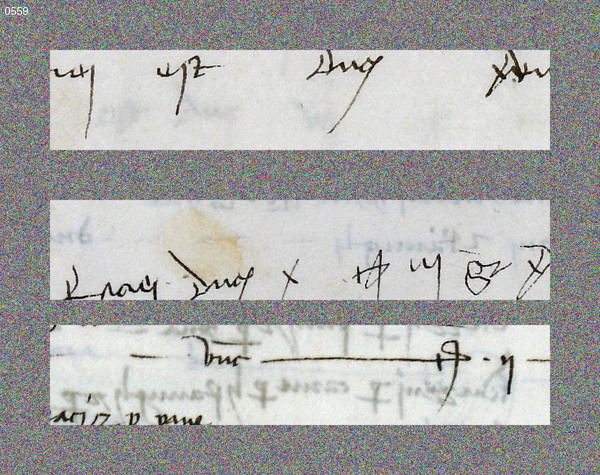} \\
     (a) & (b) & (c) \\
     \includegraphics[width=.26\textwidth]{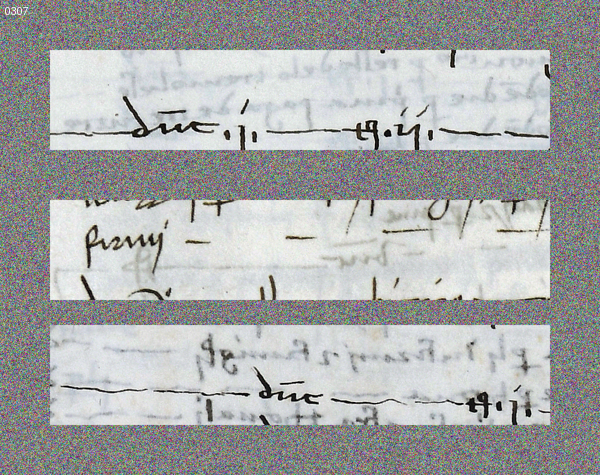} &
     \includegraphics[width=.35\textwidth]{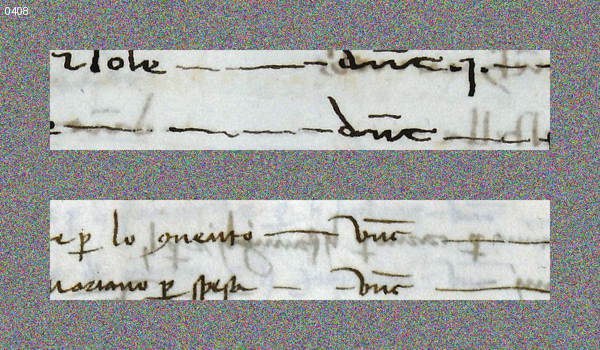} &
     \includegraphics[width=.35\textwidth]{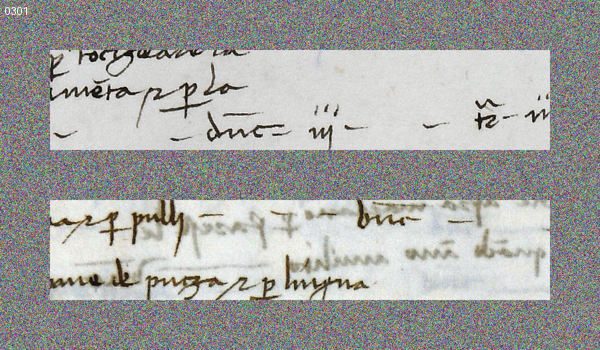} \\
     (d) & (e) & (f) \\
     
\end{tabular}
\caption{The three easiest questions in the quiz (a), (b), and (c), all answered $100\%$ correctly.
The three most misleading questions (d), (e), and (f), having misled $93.78\%$, $81.63\%$, and $77.55\%$ of participants respectively.}
\label{fig:quiz}
\end{figure}

In Fig.~\ref{fig:quiz} samples from both modalities can be seen.
Eight samples proved to mislead participants with a p-value less than $.05$, and thirteen can be described as uninformative since participants were not significantly correct.
The test was also administered to an LLM~\cite{google_gemini_1_5_pro} in PDF format; the model claimed to inspect each image individually, yet scored $51.25\%$, equivalent to random guessing, and when confronted with its results subsequently acknowledged that its responses had been fabricated rather than derived from genuine visual inspection.

\begin{figure}[b!]
\centering
\includegraphics[width=.5\textwidth]{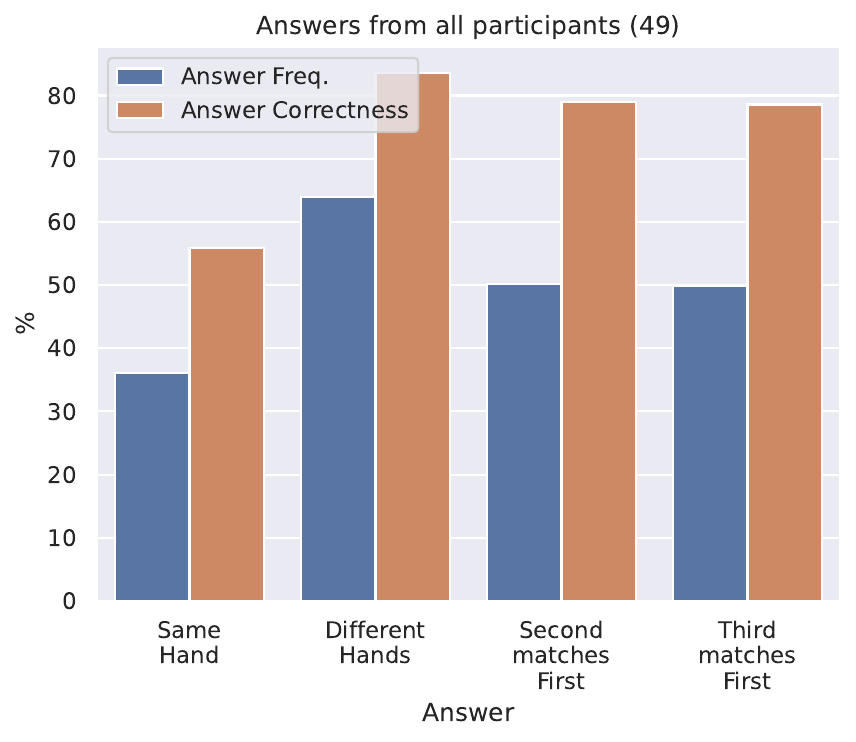}
\caption{Paleographic quiz question analysis.}
\label{fig:quiz_answers}
\end{figure}

\begin{figure}[h!]
\centering
\includegraphics[width=1.0\textwidth]{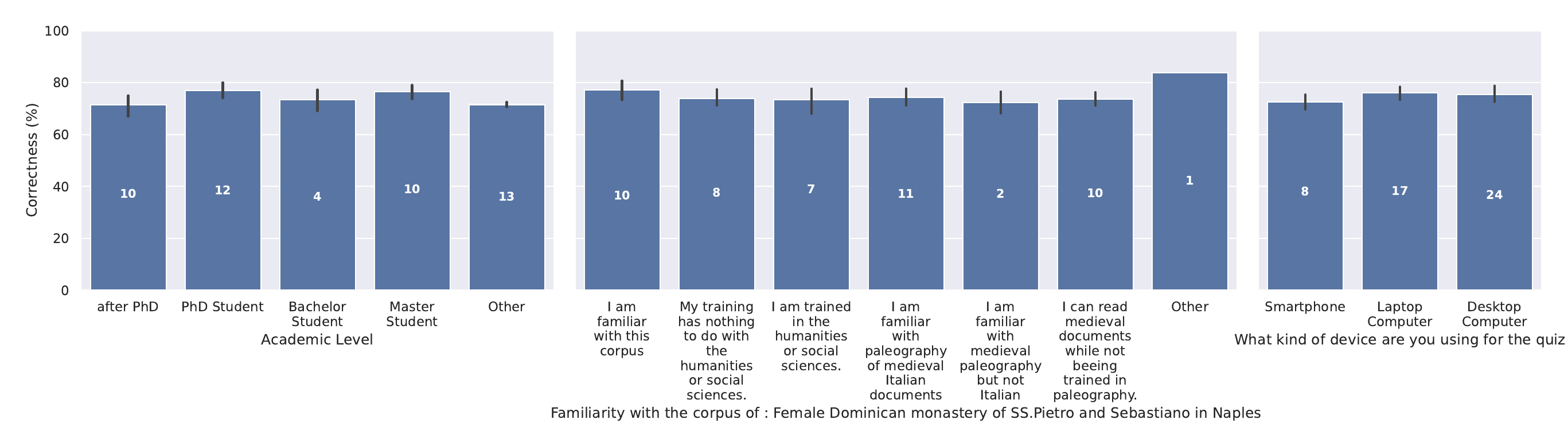}
\caption{Paleographic quiz scoring by demographic; error bars represent $95\%$ confidence intervals estimated via bootstrapping.}
\label{fig:quiz_demographics}
\end{figure}

\subsection{Zero-shot and Naive-supervision Writer Identification}
\begin{figure}[b!]
    \centering
    \begin{tabular}{ccc}
    \includegraphics[width=0.3\linewidth]{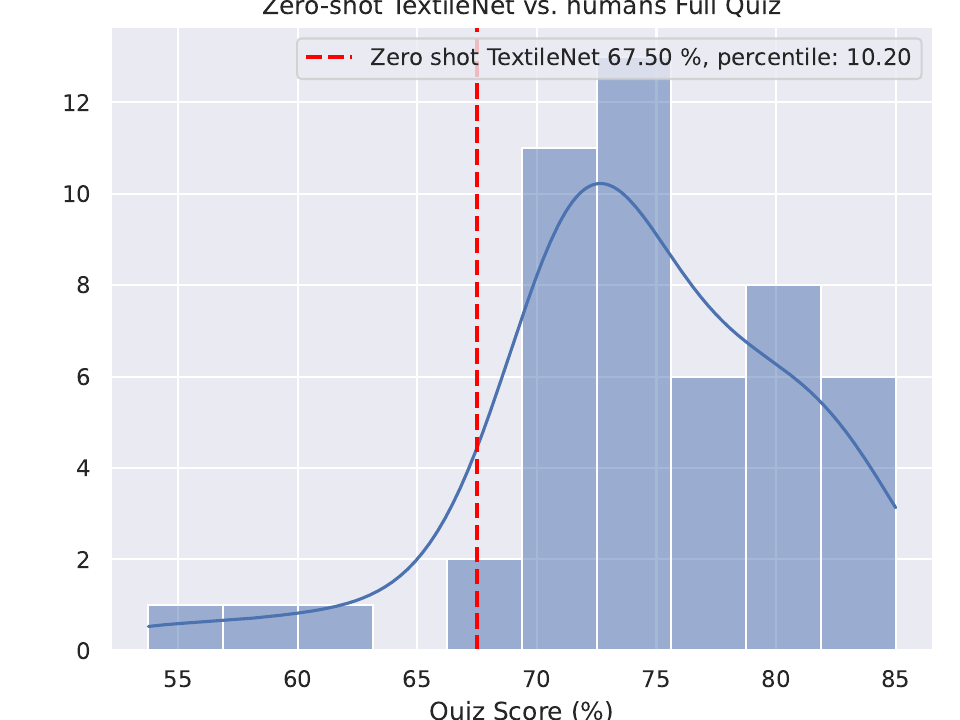} &
    \includegraphics[width=0.3\linewidth]{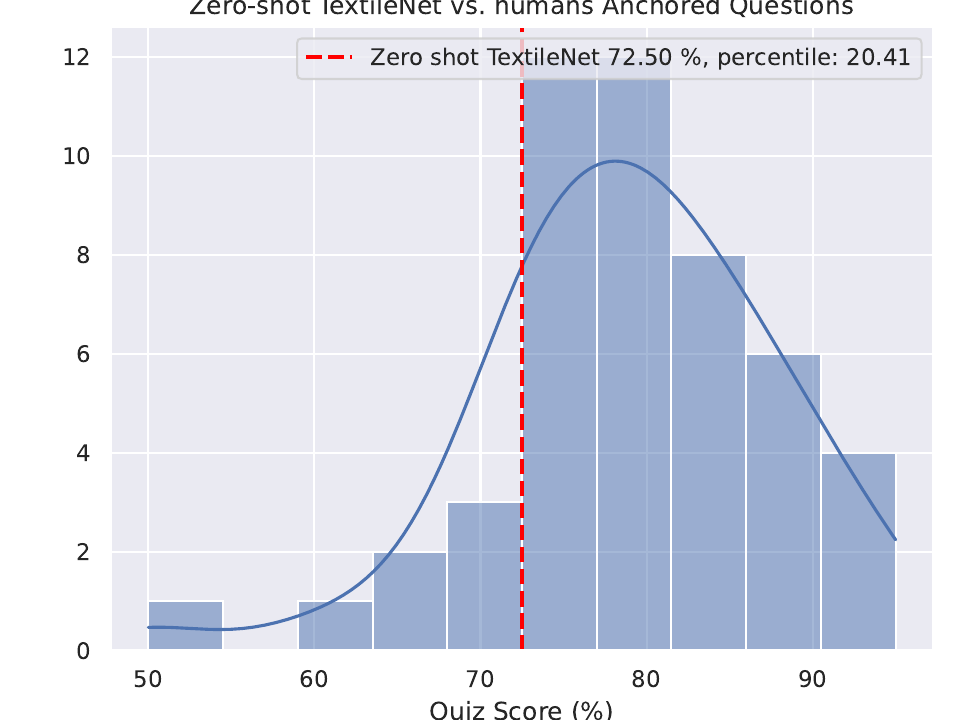} &
    \includegraphics[width=0.3\linewidth]{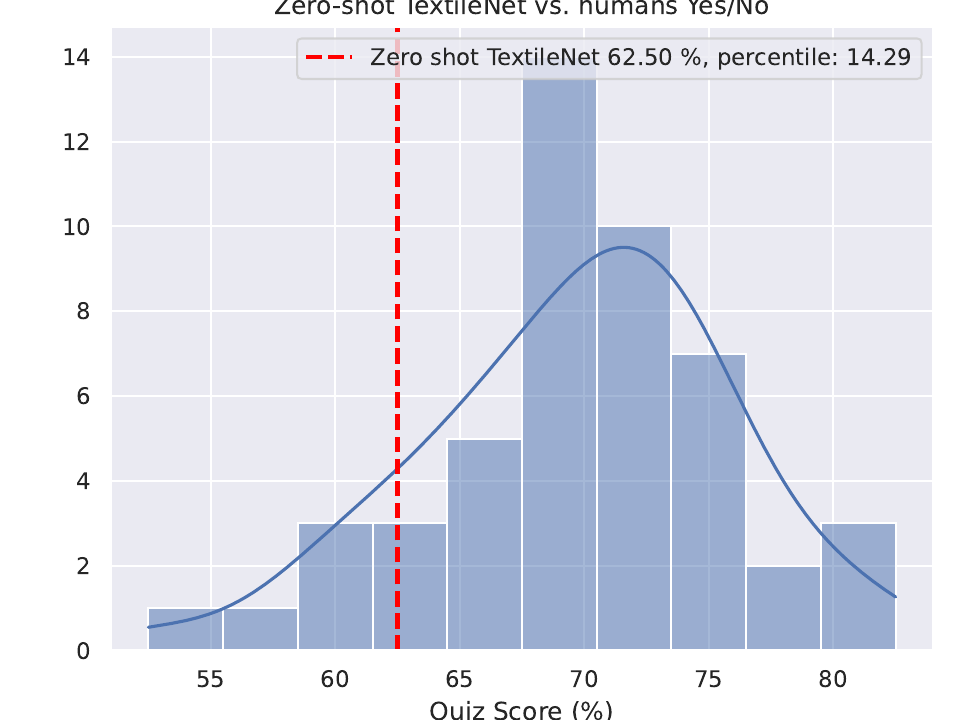}
    \end{tabular}
    \caption{Zero-shot TextileNet performance on the paleographic quiz: all questions (left), anchor/triplet questions (centre), and yes/no/pair questions (right).}
    \label{fig:autoquiz}
\end{figure}

In order to assess TextileNet's dense texture embeddings as a generic text-style understanding method, we designed an extremely naive method that uses the texture embeddings to respond to the quiz without any training.
The method operated in the following steps:
\begin{enumerate}
    \item{The quiz images were fed as wholes to TextileNet.}
    \item{Connected component analysis was performed on each sample crop using the binarization head of TextileNet.}
    \item{For each sample crop, pixel-level texture embeddings of each connected component were averaged to form a set of component embeddings, typically 10 to 50 per crop.}
    \item{For anchor (triplet) questions, a Chamfer distance~\cite{barrow1977parametric} was employed to find the nearest matching sample crop to the anchor.}
    \item{For yes/no (pair) questions, the Chamfer distance between the two crops was compared against the expected within-crop Chamfer distance estimated by random partitioning; if the between-crop distance exceeds the expected within-crop distance, the samples are deemed to come from different scribes.}
\end{enumerate}
In Fig.~\ref{fig:autoquiz} the performance of the zero-shot TextileNet on the quiz is shown, achieving $67.5\%$ overall, $72.5\%$ on anchor (triplet) questions, and $62.5\%$ on yes/no (pair) questions, compared to a $50\%$ random baseline.
The gap between the two question types is expected: yes/no questions demand an absolute identity judgment --- the respondent must internally decide whether a similarity threshold has been exceeded --- while triplet questions are a two-alternative forced choice (2AFC) requiring only a relative comparison, which sidesteps the threshold problem entirely.
Both human participants and the automatic method show the same asymmetry (Fig.~\ref{fig:autoquiz}), suggesting that future user studies of this kind should prefer the triplet framing wherever feasible.

\subsection{Zero-shot and Naive-supervision Writer Identification}
\begin{table}[t]
  \centering
  \caption{%
    Accuracy (\%) of zero-shot $k$-NN (kNN) and naive-supervision logistic-regression (LR)
    classifiers on frozen TextileNet embeddings,
    evaluated by 5-fold cross-validation on the Naples corpus.
    kNN treats classification as retrieval: no manuscript-domain training is performed
    and labels serve only to interpret the nearest-neighbour result.
    LR is included as a naive-supervision baseline, training a linear classifier on the frozen embeddings.
    \emph{Component} queries treat each connected component individually;
    \emph{Region} pools all components in one annotated region by majority vote.
    Best result per column in \textbf{bold}.
    Results are mean $\pm$ std over folds (\%).}
  \label{tab:embedding_classification}
  \begin{tabular}{l@{\hspace{1em}}cc@{\hspace{2em}}cccc}
    \toprule
    \multicolumn{3}{c}{\textbf{Method}} & \multicolumn{2}{c}{\textbf{Gender}} & \multicolumn{2}{c}{\textbf{Hand}} \\
    \cmidrule(r){1-3} \cmidrule(lr){4-5} \cmidrule(l){6-7}
    \multicolumn{1}{c}{\textbf{Classifier}} & \textbf{TextileNet} & \textbf{Positional} & \textbf{Comp.} & \textbf{Region} & \textbf{Comp.} & \textbf{Region} \\
    \midrule
    kNN & $\checkmark$ & $\times$ & $71.4 \pm 0.8$ & $79.4 \pm 6.0$ & $28.2 \pm 1.1$ & $36.5 \pm 6.7$ \\
    kNN & $\checkmark$ & $\checkmark$ & $89.7 \pm 2.3$ & $92.2 \pm 8.7$ & $35.2 \pm 2.9$ & $44.9 \pm 15.2$ \\
    kNN & $\times$ & $\checkmark$ & $\mathbf{89.9 \pm 2.3}$ & $92.2 \pm 8.7$ & $32.0 \pm 2.3$ & $46.0 \pm 8.2$ \\
    \midrule
    LR & $\checkmark$ & $\times$ & $72.7 \pm 0.7$ & $89.5 \pm 9.4$ & $37.5 \pm 0.9$ & $76.3 \pm 2.9$ \\
    LR & $\checkmark$ & $\checkmark$ & $88.2 \pm 2.2$ & $\mathbf{93.3 \pm 8.7}$ & $\mathbf{46.7 \pm 2.6}$ & $\mathbf{77.5 \pm 6.8}$ \\
    LR & $\times$ & $\checkmark$ & $87.4 \pm 2.5$ & $92.9 \pm 9.2$ & $27.3 \pm 1.8$ & $27.1 \pm 8.8$ \\
    \bottomrule
  \end{tabular}
\end{table}

Table~\ref{tab:embedding_classification} compares zero-shot kNN retrieval against a naive-supervision LR baseline, both operating on frozen TextileNet embeddings without any domain-specific fine-tuning; the results reveal a sharp asymmetry between the two classification targets.
For gender, position-only and TextileNet+Position methods achieve nearly identical accuracy ($\approx 90\%$ at component level, $\approx 92\%$ at region level), confirming that page location is the dominant discriminative cue in this corpus and substantiating the spatial confound discussed in \S\ref{sec:challenges}.
For writer identity, by contrast, the LR position-only condition collapses to $27.1\%$ at region level while LR TextileNet-only reaches $76.3\%$, demonstrating that the texture embeddings carry genuine writer-specific signal that goes well beyond page location.
We note that the logistic regression results for writer identity at region level are broadly in the same range as methods compared in~\cite{seuret2020icfhr}, though direct comparison is not possible due to differences in corpus, scribe count, and evaluation protocol.
The high variance observed in several Hand/Region cells reflects the small number of regions per fold; Component-level evaluation, which provides many more samples per fold, yields tighter confidence intervals and should be considered the more reliable estimate.
\section{Conclusion}

\subsection{Limitations}
TextileNet inference on full manuscript pages currently requires a GPU with at least 8~GB of VRAM, and training requires more than 40~GB; this makes the system impractical on consumer hardware and limits iterative experimentation.
The model currently has no mechanism to detect and discount visually unreliable regions: horizontal ruling lines, heavy bleed-through, and struck-through or corrected text all produce spurious texture activations that are indistinguishable from genuine handwriting style at the embedding level, and their influence is not suppressed during downstream classification.
The naive application of TextileNet embeddings through Chamfer-distance comparison achieves only $67.5\%$ on the paleographic quiz, indicating that while the embedding space encodes style information, more principled aggregation and comparison strategies are required before the approach can reliably serve open-set scribe discrimination.
Finally, while the dense per-pixel embeddings may in principle contain enough information to segment a page by scribal hand, a robust unsupervised method for estimating the number of distinct hands on a page is still missing, which prevents a fully automatic segmentation pipeline from being assembled from the current components.
In conclusion one could say that TextileNet appears promising for zero-shot exploratory writer-style analysis, but the current evidence is preliminary and should be validated on larger and more diverse manuscript corpora.

\subsection{Discussion}
The near-identical gender classification accuracy of position-only and TextileNet $+$\allowbreak{} Position conditions in Table~\ref{tab:embedding_classification} lends strong empirical support to the institutional interpretation advanced in \S\ref{sec:challenges}: the spatial location of writing is a near-sufficient signal for separating the labeled genders in this corpus, reflecting the structural division between supervisory male scribes and record-keeping nuns rather than any intrinsic difference in handwriting style between men and women.
The paleographic quiz results (Fig.~\ref{fig:quiz_answers}) illustrate the intrinsic difficulty of the attribution problem: participants ranging from lay historians to trained paleographers were systematically misled by eight questions and could not perform significantly above chance on a further thirteen, and a state-of-the-art multimodal large language model~\cite{google_gemini_1_5_pro} scored $51.25\%$, indistinguishable from random guessing.
The zero-shot TextileNet method achieves $67.5\%$ overall and $72.5\%$ on anchor (triplet) questions, approaching human performance on the question type most suited to fine-grained texture comparison, which suggests that dense texture embeddings encode writer-specific information at a level broadly competitive with time-constrained human judgment.
A consistent asymmetry appears across both human participants and the automatic method (Fig.~\ref{fig:autoquiz}): triplet (anchor) questions are answered more reliably than yes/no (pair) questions ($72.5\%$ vs $62.5\%$ for TextileNet).
Yes/no questions require an absolute identity judgment --- the respondent must decide whether similarity exceeds an internal threshold --- while triplet questions reduce the problem to a two-alternative forced choice (2AFC), a paradigm well-established in psychophysics as more reliable precisely because it eliminates the threshold component.
We recommend that future user studies in paleographic style comparison adopt the triplet framing as the default.
A primary practical goal of this work is to equip historians with exploratory tools rather than oracular classifiers: the random-projection RGB visualization of the texture embedding field (Fig.~\ref{fig:1401_example}) requires no labeled data and provides an immediate, interpretable summary of style variation across a page, capable of surfacing both confident attributions and contested regions for expert review.
A further concrete use case enabled by these embeddings is the automatic flagging of ambiguous or controversial annotations: the misleading quiz items identified in \S~\ref{sec:challenges} correspond precisely to the kind of borderline cases where a tool that surfaces uncertainty --- rather than forcing a binary decision --- would provide the most value to a practicing paleographer.

\subsection{Future Work}
The most immediate extension is domain adaptation of the synthesis pipeline: generating pseudo-pages that reproduce the visual properties of the target manuscript tradition --- including bleed-through, irregular baselines, age-related staining, and period-specific script morphology --- should reduce the domain gap and improve the quality of the learned embeddings on historical documents.
A more resource-efficient architecture for the TextileNet heads, such as replacing the dense cascade of $1\times1$ convolutions with lightweight factorized or attention-based alternatives, would make full-page inference feasible on consumer hardware and widen the potential user base beyond institutions with specialist GPU infrastructure.
The most consequential open problem is the development of a principled, unsupervised algorithm for decomposing the per-pixel embedding field of a page into a variable number of spatially coherent scribe regions; a hierarchical clustering approach operating directly on the embedding space, combined with an interactive visualization interface, would transform the current batch-processing analytics into a practical, session-based paleographic annotation assistant.

\end{document}